\documentclass[runningheads]{llncs}
\usepackage{makeidx}  % allows for indexgeneration
\usepackage{amssymb}
\usepackage{graphicx}
\usepackage{epsfig}
\usepackage{amsmath}
\usepackage{array}
\usepackage[T1]{fontenc}
\newcolumntype{M}[1]{>{\centering\arraybackslash}m{#1}}
\newcolumntype{L}[1]{>{\flushleft\arraybackslash}m{#1}}
\newcommand{\etal}{et al.}
\newcommand{\ie}{i.e.}
\newcommand{\eg}{e.g.}

\makeatletter
\newcommand{\printfnsymbol}[1]{%
  \textsuperscript{\@fnsymbol{#1}}%
}
\makeatother

\begin{document}
\title{Unsupervised Surgical Instrument Segmentation via Anchor Generation and Semantic Diffusion}
%in Robot-Assisted Surgery Video

%\thanks{Supported by organization x.}
%
\titlerunning{Unsupervised Surgical Instrument Segmentation}
\author{
Daochang Liu\inst{1,5}\thanks{Equal contribution} \and
Yuhui Wei\inst{1}\printfnsymbol{1} \and
Tingting Jiang\inst{1} \and
Yizhou Wang\inst{1,3,4} \and
Rulin Miao\inst{2} \and
Fei Shan\inst{2} \and
Ziyu Li\inst{2}}

% index{Liu, Daochang}
% index{Wei, Yuhui}
% index{Jiang, Tingting}
% index{Wang, Yizhou}
% index{Miao, Rulin}
% index{Shan, Fei}
% index{Li, Ziyu}

\authorrunning{D. Liu et al.}
\institute{
NELVT, Department of Computer Science, Peking University, Beijing, China \and
Peking University Cancer Hospital, Beijing, China \and
Center on Frontiers of Computing Studies, Peking University, Beijing, China \\ \and
Advanced Institute of Information Technology, Peking University, Hangzhou, China \and
Deepwise AI Lab, Beijing, China}

% Advanced Institute of Information Technology


\maketitle              % typeset the header of the contribution
\begin{abstract}

Surgical instrument segmentation is a key component in developing context-aware operating rooms.  
Existing works on this task heavily rely on the supervision of a large amount of labeled data, which involve laborious and expensive human efforts.
In contrast, a more affordable unsupervised approach is developed in this paper. 
To train our model, we first generate anchors as pseudo labels for instruments and background tissues respectively by fusing coarse handcrafted cues. 
Then a semantic diffusion loss is proposed to resolve the ambiguity in the generated anchors via the feature correlation between adjacent video frames.
In the experiments on the binary instrument segmentation task of the 2017 MICCAI EndoVis Robotic Instrument Segmentation Challenge dataset, the proposed method achieves 0.71 IoU and 0.81 Dice score without using a single manual annotation, which is promising to show the potential of unsupervised learning for surgical tool segmentation.

%practical and 
% manual annotations
% has drawn intensive research attention  active research

 %\and Pseudo label generation
\end{abstract}
\section{Introduction}

Instrument segmentation in minimally invasive surgery is fundamental for various advanced computer-aided intervention techniques such as automatic surgical skill assessment and intra-operative guidance systems~\cite{EndoVis2017}. 
Given its importance, surgical instrument segmentation has witnessed remarkable progress from early traditional methods~\cite{2014MI,2015TMI,2016MIA} to recent approaches using deep learning~\cite{2017IROS,2017MICCAI,2018ICMLA,2018MICCAI_MILLETARI,2019EMBC_HASAN,2019EMBC_NI,2019MLMI,2019MICCAI_ISLAM,2019MICCAI_JIN,2020Arxiv,2020JACS}. 
However, such success is largely built upon supervised learning from a large amount of annotated data, which are very expensive and time-consuming to collect in the medical field, especially for the segmentation task on video data.
Besides, the generalization ability of supervised methods is almost inevitably hindered by the domain gaps in real-world scenarios across different hospitals and procedure types. 
% the availability of
% labeled datasets for instrument segmentation are usually limited to small sizes~\cite{2018MICCAI-LABELS},  
%due to the annotation cost patients

In the literature, several attempts have been made to handle the lack of manual annotations~\cite{2018IJCARS,2018MICCAI-LABELS,2019ICRA,2019IJCARS,2019MICCAI_JIN}. 
Image level annotations of tool presence were utilized in~\cite{2019IJCARS,2018MICCAI-LABELS} to train neural networks in a weakly-supervised manner. 
Jin \etal~\cite{2019MICCAI_JIN} propagated the ground truth across neighboring video frames using motion flow for semi-supervised learning, while Ross \etal~\cite{2018IJCARS} reduced the number of necessary labeled images by employing re-colorization as a pre-training task. Recently, a self-supervised approach was introduced to generate labels using the kinematic signal in the robot-assisted surgery~\cite{2019ICRA}.
Compared to prior works, this study steps further to essentially eliminate the demand for manual annotations or external signals by proposing an unsupervised method for the binary segmentation of surgical tools.
Unsupervised learning has been successfully investigated in other surgical domains such as surgical workflow analysis~\cite{2017Arxiv} and surgical motion prediction~\cite{2018MICCAI_DIPIETRO}, implying its possibility in instrument segmentation.
% in the deep learning era weaken the requirement for... the lack limit the generalization  enhenced manipulation In the computer vision field, unsupervised saliency for daily objects

\begin{figure}[!pb]
\begin{center}
   \includegraphics[width=0.93\linewidth]{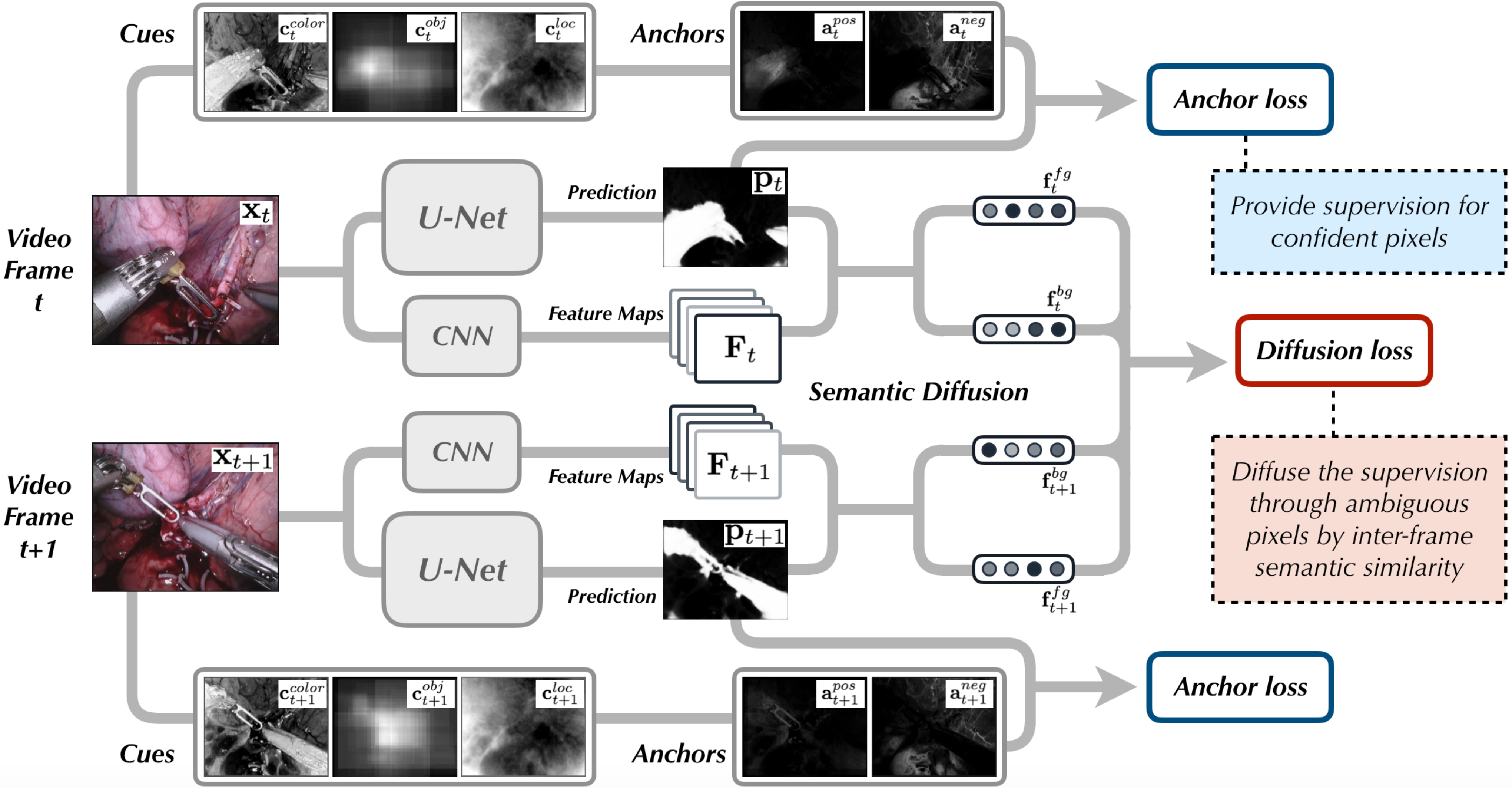}
\end{center}
   \caption{Framework overview}
\label{fig_overview}
\end{figure}

Our method, which includes \textit{anchor generation} and \textit{semantic diffusion}, learns from the general prior knowledge about surgical tools.
As for the anchor generation, we present a new perspective to train neural networks in the absence of labeled data, \ie, generating reliable pseudo labels from simple cues.
A diverse collection of cues, including color, objectness, and location cues, are fused to be positive anchors and negative anchors, which correspond to pixels highly likely to be instruments and backgrounds respectively.
Although individual cues are coarse and biased, our fusion process leverages the complementary information in these cues and thereby suppresses the noise.
The segmentation model is then trained based on these anchors.
However, since the anchors only cover a small portion of image pixels, a semantic diffusion loss is designed to propagate supervisory signals from anchor pixels to remaining pixels that are ambiguous to be instruments or not.
The core idea of this loss is to exploit the temporal correlation in surgery videos.
Specifically, adjacent video frames should share similar semantic representations in both instrument and background regions.

In the experiments on the EndoVis 2017 dataset~\cite{EndoVis2017}, the proposed method achieves encouraging results (0.71 IoU and 0.81 Dice) on the binary segmentation task without using a single manual annotation, indicating its potential to reduce cost in clinical applications.
Our method can also be easily extended to the semi-supervised setting and obtain performance comparable to the state-of-the-art. 
In addition, experiments on the ISIC 2016 dataset~\cite{ISIC16} demonstrate that the proposed model is inherently flexible to be applied in other domains like skin lesion segmentation.
In summary, our contributions are three-fold: 1) An unsupervised approach for binary segmentation of surgical tools 2) A training strategy by generating anchor supervision from coarse cues 3) A semantic diffusion loss to explore the inter-frame semantic similarity.

% instrument_dataset_1_train-frame178.png
% 0.2610970208871412
% 0.4140791970208094
% instrument_dataset_7_train-frame168.png
% 0.3687677917852786
% 0.5388317784776279
% instrument_dataset_5_train-frame072.png
% 0.6128584916794524
% 0.7599656074492801
% instrument_dataset_6_train-frame148.png
% 0.6457917714748316
% 0.7847794388911337
% instrument_dataset_3_train-frame006.png
% 0.9078450617883619
% 0.9516968437021534
% instrument_dataset_3_train-frame007.png
% 0.8941766278508237
% 0.9441322574710227
% instrument_dataset_7_train-frame037.png
% 0.8978919210616408
% 0.9461992130293478
% instrument_dataset_5_train-frame018.png
% 0.7734845902632721
% 0.8722766405863713

\section{Method}

As illustrated in Fig.~\ref{fig_overview}, our unsupervised framework\footnote{By unsupervised, we mean no manual annotation of surgical instruments is used.} consists of two aspects, 1) generating anchors to provide initial training supervision 2) augmenting the supervision by a semantic diffusion loss. 
Our framework is elaborated as follows.

\subsection{Anchor Generation}

In conventional supervised methods, human knowledge is passed to the segmentation model through annotating large-scale databases.
To be free of annotation, we encode the knowledge about surgical instruments into hand-designed cues instead and generate pseudo labels for training.
The selection of such cues should adhere to two principles, \ie, \textit{simplicity} and \textit{diversity}. The simplicity of cues prevents virtually transferring intensive efforts from the data annotation to the design process of cues, while the diversity enriches the valuable information that we can take advantage of in different cues.
Based on these principles, three cues are computed, including color, objectness and location cues.
Given a video frame ${\mathbf x}_{t} \in \mathbb{R}^{HW}$ at time $t$ with height $H$ and width $W$, probability maps ${\mathbf c}_{t}^{color} \in {[0,1]}^{HW}$, ${\mathbf c}_{t}^{obj} \in {[0,1]}^{HW}$, ${\mathbf c}_{t}^{loc} \in {[0,1]}^{HW}$ are extracted according to the three cues respectively, which are fused as pseudo labels later.

\textit{Color.} 
Color is an obvious visual characteristic to distinguish instruments from the surrounding backgrounds. 
Surgical instruments are mostly of grayish and plain colors, while the background tissues tend to be reddish and highly-saturated. 
Therefore, we multiply the inverted A channel in the LAB space, i.e., one minus the channel, and the inverted S channel in the HSV space to yield the probability map ${\mathbf c}_{t}^{color}$.

\textit{Objectness.} 
Another cue can be derived from to what extent the image region is object-like.
Surgical instruments are often with well-defined boundaries, while the background elements scatter around and fail to form a concrete shape. 
In detail, the objectness map ${\mathbf c}_{t}^{obj}$ is retrieved using a class-agnostic object detector~\cite{Objectness}.
Although this detector originally targets at daily scenes, we find it also give rich information in abdominal views.

% amorphous

\textit{Location.} 
The third cue is based on the pixel location of instrument in the screen. 
Instead of a fixed location prior, an adaptive and video-specific location probability map is obtained by averaging the color maps across the whole video: ${\mathbf c}_{t}^{loc} = \frac{1}{T} \sum_{t}{\mathbf c}_{t}^{color}$, where $T$ is the video length.
The location map roughly highlights the image areas where instruments frequently appear in this video.

\textit{Anchor Generation.}
As shown in Fig.~\ref{fig_overview}, the resultant cue maps are very coarse and noisy. 
Therefore, anchors are generated from these cues to suppress the noise.
Concretely, the positive anchor ${\mathbf a}_{t}^{pos} \in {[0,1]}^{HW}$\footnote{[0, 1] means values are between 0 and 1 both inclusively.} is defined as the element-wise product of all cues: ${\mathbf a}_{t}^{pos} = {\mathbf c}_{t}^{color}{\mathbf c}_{t}^{obj}{\mathbf c}_{t}^{loc}$, which captures the confident instrument regions that satisfy all the cues.
Similarly, the negative anchor ${\mathbf a}_{t}^{neg} \in {[0,1]}^{HW}$ is defined as the element-wise product of all inverted cues: ${\mathbf a}_{t}^{neg} = ({\mathbf 1} - {\mathbf c}_{t}^{color})({\mathbf 1} - {\mathbf c}_{t}^{obj})({\mathbf 1} - {\mathbf c}_{t}^{loc})$, which captures the confident background regions that satisfy none of the cues.
As in Fig.~\ref{fig_overview}, the false response is considerably minimized in the generated anchors.
% Intuitively, the errors in the coarse cues comes

\textit{Anchor Loss.}
The anchors are then regarded as pseudo labels to train the segmentation network, a vanilla U-Net~\cite{Unet} in this paper.
We propose an anchor loss to encourage network activation on the positive anchor and inhibit activation on the negative anchor:
\begin{equation}
{\textstyle \mathcal{L}_{anc}({\mathbf x}_{t}) = \frac{1}{HW}\sum_{i} - {\mathbf a}^{pos}_{t,i}{\mathbf p}_{t,i} - {\mathbf a}^{neg}_{t,i}(1 - {\mathbf p}_{t,i})}
\end{equation}
where ${\mathbf p}_{t} \in {[0,1]}^{HW}$ denotes the prediction map from the network and $i$ is the pixel index. 
The loss is computed for each pixel and averaged over the whole image. 
Compared to the standard binary cross-entropy, this anchor loss only imposes supervision on the pixels that are confident to be instruments or backgrounds, keeping the network away from being disrupted by the noisy cues.
However, the anchors only amount to a minority of image pixels. 
On the remaining ambiguous pixels outside the anchors, the network is not supervised and its behavior is undefined.
Such a problem is tackled by the following semantic diffusion loss.

\subsection{Semantic Diffusion}
Apart from the cues mentioned above, temporal coherence is another natural source of knowledge for unsupervised learning in the sequential data.
We argue that the instruments in adjacent video frames usually share similar semantics, termed as \textit{inter-frame instrument-instrument similarity}.
This temporal similarity is assumed to be stronger than the semantic similarity between the instrument and the background within a single frame, \ie, the \textit{intra-frame instrument-background similarity}.
To this end, the semantic feature maps from a pre-trained convolutional neural network (CNN) are first aggregated within the instrument and background regions respectively using the prediction map:
\begin{equation}
{\textstyle {\mathbf f}^{fg}_{t} = \sum_{i} {\mathbf p}_{t,i}{\mathbf F}_{t,i}
 \;\;\;\;\;\;\;\;
{\mathbf f}^{bg}_{t} = \sum_{i} (1 - {\mathbf p}_{t,i}){\mathbf F}_{t,i}}
\end{equation}
where ${\mathbf F}_{t} \in \mathbb{R}^{HW \times D}$ represents the CNN feature maps of frame ${\mathbf x}_{t}$, and ${\mathbf F}_{t,i} \in \mathbb{R}^{D}$ denotes the features at pixel $i$, and $D$ is the channel number, ${\mathbf f}^{fg}_{t} \in \mathbb{R}^{D}$ and ${\mathbf f}^{bg}_{t} \in \mathbb{R}^{D}$ are the aggregated features for the instrument and the background correspondingly.
Then given two adjacent frames ${\mathbf x}_{t}$ and ${\mathbf x}_{t+1}$, a semantic diffusion loss in a quadruplet form is proposed to constrain the \textit{inter-frame instrument-instrument similarity} to be higher than the \textit{intra-frame instrument-background similarities} by a margin:
\begin{equation}
 \mathcal{L}_{dif}^{fg}({\mathbf x}_{t}, {\mathbf x}_{t+1}) = \mathrm{max}(\phi({\mathbf f}^{fg}_{t}, {\mathbf f}^{bg}_{t}) + \phi({\mathbf f}^{fg}_{t+1}, {\mathbf f}^{bg}_{t+1}) - 2 \phi({\mathbf f}^{fg}_{t},{\mathbf f}^{fg}_{t+1}) + m^{fg}, 0)
\end{equation}
where $\phi(\cdot,\cdot)$ denotes the cosine similarity between two features and $m^{fg}$ is a hyperparameter controlling the margin. 
Likewise, another semantic diffusion loss can be formulated to enforce the \textit{inter-frame background-background similarity}:
\begin{equation}
 \mathcal{L}_{dif}^{bg}({\mathbf x}_{t}, {\mathbf x}_{t+1}) = \mathrm{max}(\phi({\mathbf f}^{fg}_{t}, {\mathbf f}^{bg}_{t}) + \phi({\mathbf f}^{fg}_{t+1}, {\mathbf f}^{bg}_{t+1}) - 2 \phi({\mathbf f}^{bg}_{t},{\mathbf f}^{bg}_{t+1}) + m^{bg}, 0).
\end{equation}
Lastly, the anchor loss and the semantic diffusion loss are optimized collectively:
\begin{equation}
 \mathcal{L}_{full}({\mathbf x}_{t}, {\mathbf x}_{t+1}) = \mathcal{L}_{anc}({\mathbf x}_{t}) + \mathcal{L}_{anc}({\mathbf x}_{t+1}) + \mathcal{L}_{dif}^{fg}({\mathbf x}_{t}, {\mathbf x}_{t+1}) + \mathcal{L}_{dif}^{bg}({\mathbf x}_{t}, {\mathbf x}_{t+1}).
\end{equation}
Driven by the semantic diffusion loss, the initial signals on the confident anchor pixels are propagated to remaining ambiguous pixels.
Our network benefits from such augmented supervision and outputs accurate and complete segmentation.
Note that the semantic diffusion loss is generally not restricted to adjacent frames and can be also imposed on any image pair exhibiting inter-image similarity.

\section{Experiment}

\noindent
\textit{Dataset.}
Our method is evaluated on the dataset of the 2017 MICCAI EndoVis Robotic Instrument Segmentation Challenge~\cite{EndoVis2017} (EndoVis 2017), which consists of 10 abdominal porcine procedures videotaped by the da Vinci Xi systems.
Our work focuses on the binary instrument segmentation task, where each frame is separated into instruments and backgrounds. 
As our method is unsupervised, we do not use any annotations during the training process.
Note that the ground truth of the test set is still held out by the challenge organizer.

\noindent
\textit{Setup.}
Experiments are carried out in two different settings.
1) \textit{Train Test (TT)}: This setting is common for supervised methods, where the learning and the inference are performed on two different sets of data. 
In this setting, we follow the previous convention~\cite{2019MICCAI_JIN} and conduct 4-fold cross-validation on the released 8 training videos of EndoVis 2017, with the same splits as prior works.
Our method can attain real-time online inference speed in this setting.
2) \textit{Single Stage (SS)}: This is a specific setting for our unsupervised method.
Since the learning involves no annotation, we can directly place the learning and the inference on the same set of data, \ie, the released training set of EndoVis 2017.
In application scenarios, the model needs to be re-trained when new unseen data comes, therefore this setting is more suitable for the offline batch analysis. 
Following previous work~\cite{2019MICCAI_JIN}, we use intersection-over-union (IoU) and Dice coefficient to measure our performance.

\noindent
\textit{Implementation Details.} % check later
We extract the semantic feature maps from the $relu5\_3$ layer of the VGG16~\cite{Vgg} pre-trained on ImageNet, which are interpolated to the same size as the prediction map. 
The VGG16 extractor is frozen when training our U-Net.
The margin factors $m^{fg}$ and $m^{bg}$ are set as $0.2$ and $0.8$.
The prediction map is thresholded to be final segmentation mask using the Otsu algorithm~\cite{Otsu}. Our implementation uses official pre-trained CNN models and parameters in PyTorch~\cite{PyTorch}. Codes will be released to offer all details.

\subsection{Results on EndoVis 2017}

Results on EndoVis 2017 are reported in Table~\ref{table_1}. Firstly we assess the network performance only using the anchor loss $\mathcal{L}_{anc}$ based on our cues, where we get the basic performance. 
After we combine the semantic diffusion losses, especially the background semantic diffusion loss $\mathcal{L}_{dif}^{bg}$, the performance is strikingly improved. 
This result proves our assumption that adjacent video frames are similar to each other in both foreground and background regions. 
Since the background area is relatively more similar between the video frames, it is seen from Table~\ref{table_1} that $\mathcal{L}_{dif}^{bg}$ brings more improvement on the performance than $\mathcal{L}_{dif}^{fg}$.

\begin{table}[t]
\begin{center}
\caption{Results of the binary segmentation task from EndoVis 2017. Experimental results in the setting SS are reported.}
\label{table_1}
\begin{tabular}{M{0.10\linewidth} M{0.10\linewidth} M{0.10\linewidth} |M{0.19\linewidth}|M{0.19\linewidth}}
\hline
$\mathcal{L}_{anc}$ & $\mathcal{L}_{dif}^{fg}$ & $\mathcal{L}_{dif}^{bg}$ & \textbf{IoU (\%)} & \textbf{Dice (\%)} \\
\hline
 \checkmark & & & 49.47 & 64.21 \\
 \checkmark & \checkmark & & 50.78 & 65.16 \\
 \checkmark & & \checkmark & 67.26 & 78.94 \\
 \checkmark & \checkmark & \checkmark & \textbf{70.56} & \textbf{81.15}\\
\hline
\end{tabular}
\end{center}
\end{table}

\subsection{The Choice of Cues}

Different combinations of cues are examined to research their effects on the network performance. 
Here we run the Otsu thresholding algorithm~\cite{Otsu} not only on the network prediction map ${\mathbf p}$ but also on the corresponding positive anchor ${\mathbf a}^{pos}$ and the inverted negative anchor ${\mathbf 1}-{\mathbf a}^{neg}$ to generate segmentation masks.
The Otsu algorithm is adaptive to the disparate intensity level of the probabilistic maps.
The resultant masks are then evaluated against the ground truth.
As the results shown in Table~\ref{table_2}, the best network performance comes from the combination of all three cues, because more kinds of cues can provide extra information from different aspects.
Meanwhile, a single kind of cue may produce good results on the anchors, but may not be helpful to the final network prediction, because a single cue may contain lots of noise and it needs to be filtered out by the fusion with other useful cues.
Also, different kinds of cues have varying effects on the network performance.
For example, it is noticed that the $\textbf{c}^{color}$ and $\textbf{c}^{obj}$ cues are more important than the $\textbf{c}^{loc}$ from the table.

\begin{table}[t]
\begin{center}
\caption{The choice of cues (Setting SS)}
\label{table_2}
\begin{tabular}{M{0.08\linewidth} M{0.08\linewidth} M{0.08\linewidth}|M{0.11\linewidth} M{0.11\linewidth} M{0.11\linewidth}|M{0.11\linewidth} M{0.11\linewidth} M{0.11\linewidth}}
\hline
 & & & \multicolumn{3}{c|}{\textbf{IoU (\%)}} & \multicolumn{3}{c}{\textbf{Dice (\%)}}\\
${\mathbf c}^{color}$ & ${\mathbf c}^{obj}$ & ${\mathbf c}^{loc}$ & ${\mathbf a}^{pos}$ & ${\mathbf 1}-{\mathbf a}^{neg}$ & ${\mathbf p}$ & ${\mathbf a}^{pos}$ & ${\mathbf 1}-{\mathbf a}^{neg}$ & ${\mathbf p}$ \\
\hline
 \checkmark & & & 55.60 & 55.60 & 45.27 & 69.21 & 69.21 & 60.09 \\
 & \checkmark & & 16.01 & 16.01 & 14.23 & 26.57 & 26.57 & 23.57 \\
 & & \checkmark & 16.90 & 16.90 & 21.32 & 28.11 & 28.11 & 33.97 \\
 & \checkmark & \checkmark & 20.28 & 18.93 & 47.39 & 32.48 & 30.99 & 62.51 \\
 \checkmark & & \checkmark & 41.44 & 19.21 & 43.74 & 57.00 & 31.46 & 59.30 \\
 \checkmark & \checkmark & & 38.69 & 22.09 & 63.27 & 53.70 & 35.19 & 75.56 \\
 \checkmark & \checkmark & \checkmark & 38.64 & 18.53 & \textbf{70.56} & 53.73 & 30.56 & \textbf{81.15} \\
\hline
\end{tabular}
\end{center}
\end{table}

\subsection{Compared to Supervised Methods}

At present, unsupervised instrument segmentation is still less explored, with few methods that can be directly compared with.
Therefore, to provide an indirect reference, our method is adjusted to the semi-supervised and fully-supervised settings and compared with previous supervised methods in Table~\ref{table_3}.
When fully-supervised, we substitute the anchors by the ground truth on all the frames. 
Since our contribution is not in the network architecture and we do not use special modules such as attention beyond the U-Net, our fully-supervised performance is close to some earlier works, which can be thought of as an upper bound of our unsupervised solution.
When semi-supervised, the anchors are replaced with the ground truth on 50\% frames in the same periodical way as in~\cite{2019MICCAI_JIN}.
Our method has competitive performance with the state-of-the-art in the semi-supervised setting. 
Lastly, in the last two rows without using any annotation, we achieve the preeminent performance.
More data is exploited for learning in the setting SS than in the setting TT, which explains why the setting SS has better results. 

\begin{table}[t]
\begin{center}
\caption{Comparison with supervised methods (mean$\pm$std). Results of prior works are quoted from~\cite{2019MICCAI_JIN}. Not all the existing fully-supervised methods are listed due to limited space. Our network architecture is the vanilla U-Net.}
\label{table_3}
\begin{tabular}{M{0.16\linewidth}|M{0.28\linewidth}|M{0.11\linewidth}|M{0.19\linewidth}|M{0.19\linewidth}}
\hline
\hline
\textbf{Supervision} & \textbf{Method} & \textbf{Setting} & \textbf{IoU (\%)} & \textbf{Dice (\%)} \\
\hline
\hline
100\% & U-Net~\cite{Unet} & TT & 75.44$\pm$18.18 & 84.37$\pm$14.58 \\
100\% & Ours ($\mathcal{L}_{full}$) & TT & 81.55$\pm$14.52 & 88.83$\pm$11.50 \\
100\% & TernausNet~\cite{2018ICMLA} & TT & 83.60$\pm$15.83 & 90.01$\pm$12.50 \\
% 100\% & U-NetPlus & TT & 83.75 $\pm$ 15.36 & 90.19 $\pm$ 11.77 \\
100\% & MF-TAPNet~\cite{2019MICCAI_JIN} & TT & 87.56$\pm$16.24 & 93.37$\pm$12.93 \\
\hline
\hline
50\% & Semi-MF-TAPNet~\cite{2019MICCAI_JIN} & TT & 80.03$\pm$16.87 & \textbf{88.07}$\pm$ 13.15\\
50\% & Ours ($\mathcal{L}_{full}$) & TT & \textbf{80.33}$\pm$14.69 & 87.94$\pm$11.53 \\
% 25\% & Ours ($\mathcal{L}_{full}$) & TT & 78.38 $\pm$ 11.91 & 86.75 $\pm$ 8.29 \\
\hline
\hline
0\% & Ours ($\mathcal{L}_{full}$) & TT & 67.85$\pm$15.94 & 79.42$\pm$13.59 \\
0\% & Ours ($\mathcal{L}_{full}$) & SS & \textbf{70.56}$\pm$16.09 & \textbf{81.15}$\pm$13.79\\
\hline
\hline
\end{tabular}
\end{center}
\end{table}

\subsection{Qualitative Result}

\begin{figure}[t]
\centering
\includegraphics[width=1\linewidth]{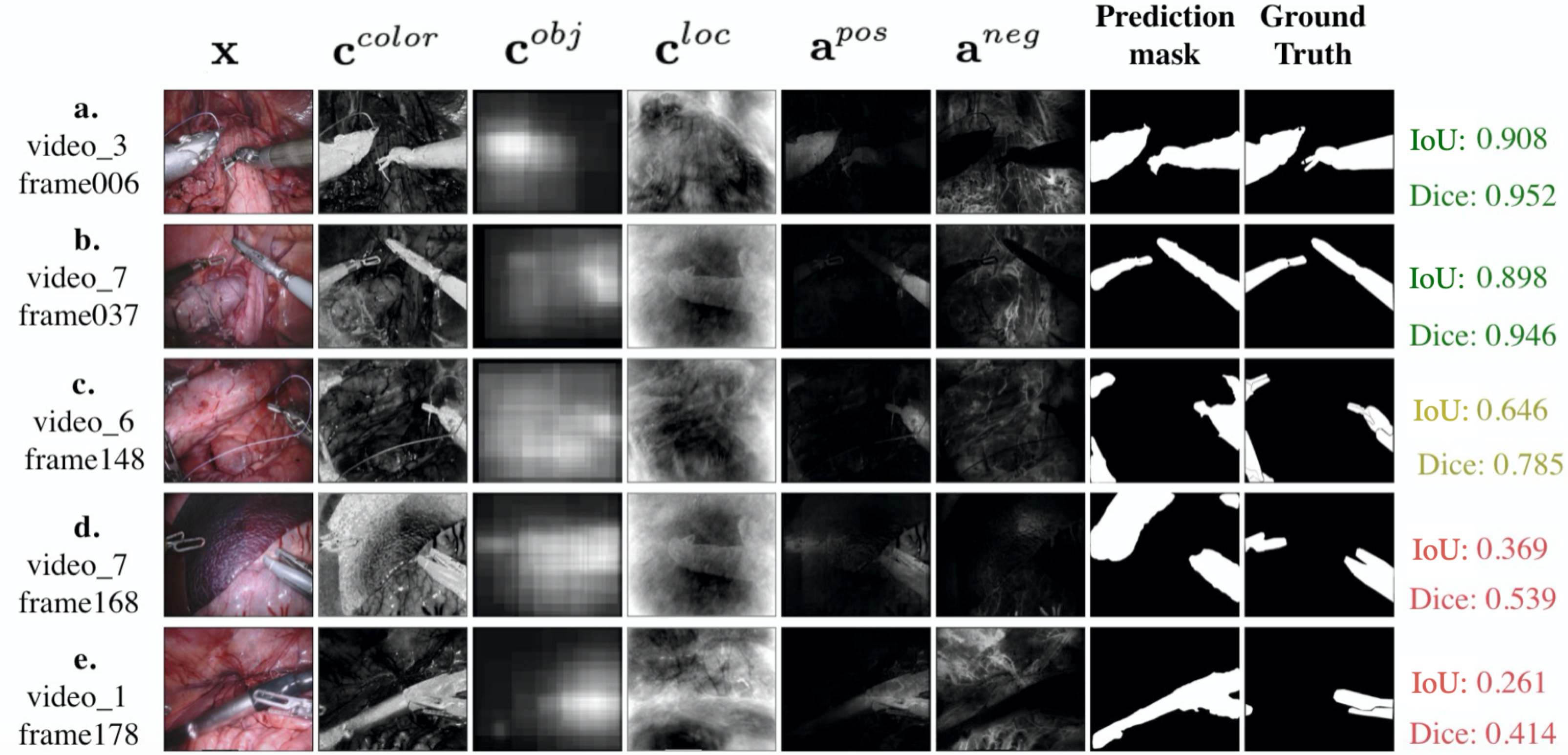}
\caption{Visual results for success and failure cases.}
\label{fig_result}
\end{figure}

In this section, some visual results from our method are plotted. 
Firstly, as seen in Fig.~\ref{fig_result}, the three cues are very coarse, \eg, the background can still be found on the color cue maps.
By the fusion of noisy cues, the anchors become purer, which are nonetheless very sparse. 
Then via the semantic diffusion loss, which augments the signals on the anchors, the network can find the lost instrument region in those anchor pictures, as shown in the success cases in Fig.~\ref{fig_result}a and Fig.~\ref{fig_result}b.
Although in some pictures there are difficulties such as complicated scenes and lighting inconstancy, we can also get good performance in these cases. 
However, there are still some failure cases, such as the special probe (Fig.~\ref{fig_result}e) that is not thought of as the instrument in the ground truth. 
Also, the dimmed light and the dark organ (Fig.~\ref{fig_result}d) can also have negative effects on the reliability of cues.
A video demo is attached in the supplementary material.

\subsection{Extension to Other Domain}

An exploratory experiment is conducted on the skin lesion segmentation task of ISIC 2016 benchmark~\cite{ISIC16} to inspect whether our model can be migrated to other domains.
We conform to the official train-test split.
Due to the dramatic color variations of lesions, the color cue is excluded.
The location cue is set as a fixed 2D Gaussian center prior since ISIC is not a video dataset.
In view of the background similarity shared by most images, we sample random image pairs for semantic diffusion.
Our flexibility is provisionally supported by the results in Table~\ref{table_4}.
Specific cues for skin lesions can be designed in future for better results.

\begin{table}[t]
\begin{center}
\caption{Results on the skin lesion segmentation task of ISIC 2016}
\label{table_4}
\begin{tabular}{M{0.17\linewidth}|M{0.26\linewidth}|M{0.12\linewidth}|M{0.19\linewidth}|M{0.19\linewidth}}
\hline
\textbf{Supervision} & \textbf{Method} & \textbf{Setting} & \textbf{IoU (\%)} & \textbf{Dice (\%)} \\
\hline
100\% & Challenge Winner~\cite{ISIC16} & TT & 84.3 & 91.0 \\
100\% & Ours ($\mathcal{L}_{full}$) & TT & 83.6 & 90.3 \\
\hline
50\% & Ours ($\mathcal{L}_{full}$) & TT & 81.1 & 88.6 \\
\hline
0\% & Ours ($\mathcal{L}_{full}$) & TT & 63.3 & 74.9 \\
0\% & Ours ($\mathcal{L}_{full}$) & SS & \textbf{64.4} & \textbf{75.7} \\
\hline
\end{tabular}
\end{center}
\end{table}

\section{Conclusion and Future Work}
This work proposes an unsupervised surgical instrument segmentation method via anchor generation and semantic diffusion, whose efficacy and flexibility are validated by empirical results.
The current framework is still limited to binary segmentation. 
In future works, multiple class-specific anchors could be generated for multi-class segmentation, while additional grouping strategies could be incorporated as post-processing to support instance or part segmentation.

\subsection*{Acknowledgments} This work was partially supported by MOST-2018AAA0102004 and the Natural Science Foundation of China under contracts 61572042, 61527804, 61625201. We also acknowledge the Clinical Medicine Plus X-Young Scholars Project, and High-Performance Computing Platform of Peking University for providing computational resources.
Thank Boshuo Wang for making the video demo.

% come from differencn sources,
% more reliable 
% false response
% to distinguish instruments from backgrounds
% and the domain knowledge
% sources of error
% pixel-labeling
% seed growing

% Compare to self-supervised pre-training or baselines?
% REPRODUCIBLE RESEARCH

% ${\mathbf x}_{t} \in \mathbb{R}^{HW}$
% ${\mathbf p}_{t} \in {[0,1]}^{HW}$
% $\hat{{\mathbf p}}_{t} \in {\{0,1\}}^{HW}$
% ${\mathbf y}_{t} \in {\{0,1\}}^{HW}$
% ${\mathbf c}_{t}^{color} \in {[0,1]}^{HW}$
% ${\mathbf c}_{t}^{obj} \in {[0,1]}^{HW}$
% ${\mathbf c}_{t}^{loc} \in {[0,1]}^{HW}$
% ${\mathbf a}_{t}^{pos} \in {[0,1]}^{HW}$
% ${\mathbf a}_{t}^{neg} \in {[0,1]}^{HW}$
% ${\mathbf F}_{t} \in \mathbb{R}^{HW \times 512}$
% ${\mathbf f}^{fg}_{t} \in \mathbb{R}^{512}$
% ${\mathbf f}^{bg}_{t} \in \mathbb{R}^{512}$
% ${\mathbf F}_{t,i}$
% ${\mathbf p}_{t,i}$

% ${\mathbf x}_{t+1} \in \mathbb{R}^{HW}$
% ${\mathbf p}_{t+1} \in {[0,1]}^{HW}$
% $\hat{{\mathbf p}}_{t+1} \in {\{0,1\}}^{HW}$
% ${\mathbf y}_{t+1} \in {\{0,1\}}^{HW}$
% ${\mathbf c}_{t+1}^{color} \in {[0,1]}^{HW}$
% ${\mathbf c}_{t+1}^{obj} \in {[0,1]}^{HW}$
% ${\mathbf c}_{t+1}^{loc} \in {[0,1]}^{HW}$
% ${\mathbf a}_{t+1}^{pos} \in {[0,1]}^{HW}$
% ${\mathbf a}_{t+1}^{neg} \in {[0,1]}^{HW}$
% ${\mathbf F}_{t+1} \in \mathbb{R}^{HW \times 512}$
% ${\mathbf f}^{fg}_{t+1} \in \mathbb{R}^{512}$
% ${\mathbf f}^{bg}_{t+1} \in \mathbb{R}^{512}$
% ${\mathbf F}_{t+1,i}$
% ${\mathbf p}_{t+1,i}$

% {\bf Acknowledgement.} 

{%\small
\bibliographystyle{splncs04}
\bibliography{mybib}
}

\end{document}